\documentclass[conference]{IEEEtran}

\usepackage{caption}
\usepackage{graphicx}
\usepackage{tabularx} 
\usepackage[normalem]{ulem}
\usepackage{makecell} 
\usepackage{xspace}
\usepackage{booktabs}
\usepackage{multirow}
\usepackage{lipsum}

\usepackage{enumitem}
\usepackage{amsmath}
\usepackage{hyperref}
\usepackage{cleveref} 
\usepackage{float}
\usepackage{adjustbox}
\setlist[itemize]{noitemsep, topsep=0pt}
\usepackage{adjustbox}
\usepackage{wrapfig}
\usepackage{subcaption}
\usepackage[normalem]{ulem}
\usepackage{algorithmicx}
\usepackage{algorithm}
\usepackage{algpseudocode} 
\usepackage{amsmath}       
\usepackage{booktabs}   
\usepackage{pifont}     
\usepackage{xcolor}     
\usepackage{colortbl}   
\usepackage{graphicx}   

\usepackage{amssymb}

\definecolor{Gray}{gray}{0.9} 

\usepackage{soul}
\hypersetup{
    colorlinks=true,
    linkcolor=blue,
    filecolor=magenta,
    urlcolor=cyan,
}

\newcommand{\system}{\textit{REIS}\xspace}

\title{On-Device Robotic Planning: Eliminating Inference Redundancy for Efficient Decision-Making}

\author{
  \IEEEauthorblockN{
    Joonhee Lee\textsuperscript{*,}\IEEEauthorrefmark{1},
    Sungmin Lee\textsuperscript{*,}\IEEEauthorrefmark{1},
    Hyunsung Shin\IEEEauthorrefmark{1},
    Hyunmi Kim\IEEEauthorrefmark{2},
    Pei Zhang\IEEEauthorrefmark{3}, and
    Jeonggil Ko\IEEEauthorrefmark{1}
    \thanks{*Both authors contributed equally to this work.}
  }
  \IEEEauthorblockA{
    \IEEEauthorrefmark{1}Yonsei University, Seoul, South Korea\\
    \{neo81389, i.am.sungmin.lee, jay040730, jeonggil.ko\}@yonsei.ac.kr
  }
  \IEEEauthorblockA{
    \IEEEauthorrefmark{2}Department of Hyperscale AI SoC Research Section, ETRI, South Korea\\
    chaos0218@etri.re.kr
  }
  \IEEEauthorblockA{
    \IEEEauthorrefmark{3}University of Michigan, Ann Arbor, MI, USA\\
    peizhang@umich.edu
  }
}

\begin{document}
\maketitle

\bstctlcite{IEEEexample:BSTcontrol}

\begin{abstract}
    Reasoning-based robotic policies using large language and vision-language models achieve strong semantic planning capabilities but mostly suffer from a high inference latency that limits practical real-time deployment. In this work, we observe that robotic reasoning workloads contain substantial temporal redundancy, where consecutive observations frequently produce identical actions and subgoals. Based on this insight, we present \system{}, a human cognition inspired robotic decision-making framework that minimizes unnecessary reasoning while preserving semantic adaptability. \system{} combines lightweight scene gating, KV-steered affordance routing, and deliberative reasoning to accelerate robotic control under embodied constraints. Experiments on ALFRED, and real-world robotic tasks demonstrate that \system{} significantly suppresses reasoning overhead while maintaining competitive task performance.
\end{abstract}


\section{Introduction}
\label{sec:Intro}

Recent advances in large-scale foundation models have enabled a ``reason-and-act'' (ReAct) paradigm for robotic decision-making, where policies use intermediate reasoning to interpret context, decompose instructions, and plan actions before execution, improving performance in complex instruction-driven navigation and manipulation tasks~\cite{zhao2025cotvla, wen2025diffusionvla}.
However, this benefit comes with substantial inference overhead. In real-world robotics, reasoning may need to be repeatedly invoked as new observations arrive, making its computational cost difficult to sustain under the limited compute, energy, and thermal budgets of onboard platforms~\cite{yin2025deepthinkvla}. This creates a fundamental tension between rich reasoning and continuous, low-latency robotic operation.

Prior work has approached this efficiency challenge from multiple directions. Model-level techniques reduce the cost of individual reasoning calls~\cite{yang2025efficientvla}, while framework-level designs separate expensive high-level LM planning from lower-level execution so that a high-level plan can guide multiple actions~\cite{song2022llmplanner,shah22lmnav}. More recent systems further improve runtime efficiency through multi-rate execution and temporal reuse of high-level information~\cite{zhang2025hirt,duan2025fastecot}. These approaches commonly rely on a per-input \textit{reasoning decision} to determine whether the current high-level reasoning remains sufficient or should be refreshed. Since this decision directly governs reasoning frequency and end-to-end latency, improving its efficiency remains an important but comparatively underexplored optimization target.

To address this gap, we propose \system{}, a reasoning-efficient ReAct framework that structures the \textit{reasoning decision} (i.e., fresh reasoning invoking) as a two-stage cascaded process. The key idea is that not all observations require new decisions.
Thus, the first stage exploits lightweight structural evidence to detect task-relevant scene changes, while the second uses richer semantic evidence to assess whether the current high-level task decision should be revised. If neither stage can confidently support the continuity of the current decision, \system{} invokes a full deliberative reasoning process. By doing so, \system{} reduces both unnecessary reasoning invocations and the cost of reasoning decision making itself.

The first stage of \system{} requires a lightweight yet robust approach to distinguish task-relevant scene changes from visual variation that should not trigger deeper reasoning. For instance, slight movement can shift the background during navigation and robotic arm movements themselves can create superficial pixel motion during manipulation. These factors challenge the scene understanding since ego-motion alters consecutive views without changing task states, with the robot's motion itself dominating the frame dynamics and obscuring meaningful changes. \system{} addresses this through an Ego-Motion Aware Head Selective Vision Similarity (EMA-HSVS) operation, which monitors selected transformer heads to isolate decision-critical structural changes from superficial fluctuations and applies \textit{Self-Mask} to exclude robot-occupied regions. This enables low-cost \textit{structural verification} that filters redundant observations before richer semantic processing.

Designing the second stage, which is invoked when structural verification detects a meaningful scene change, requires richer task-level understanding without immediately paying the cost of full multi-step reasoning. A genuine scene change does not necessarily require the current subgoal or semantic action to change; thus, the system must determine whether the detected change actually affects the robot's ongoing execution. \system{} addresses this requirement through a \textit{KV-steered Affordance Router}, which injects task-specific reasoning information prepared offline and directly evaluates subgoal progress and action validity without regenerating a full reasoning trace. This provides lower-cost \textit{semantic verification}, while unresolved cases proceed to full reasoning.

Extensive evaluations on ALFRED and real-world robotic deployments show that \system{} substantially improves execution efficiency while preserving task performance. Compared with periodic reasoning baselines, \system{} achieves an average $15.09\times$ execution speedup with only 3.73\% accuracy reduction. Reducing unnecessary reasoning also improves responsiveness on constrained edge hardware, yielding a $2.21\times$ speedup in failure detection during physical manipulation.

The contributions of this work are summarized as follows:
\begin{enumerate}[label={(\arabic*)}, leftmargin=*]

\item We introduce \system{}, a reasoning-efficient ReAct framework that treats the \textit{reasoning decision}, whether fresh reasoning is required, as a latency-optimization target and resolves it through two-stage verification.

\item We develop a lightweight \textit{structural verification} approach using \textit{EMA-HSVS} with \textit{Self-Mask} to distinguish task-relevant scene changes from superficial visual variation while excluding robot-occupied regions that can obscure meaningful changes.

\item We develop \textit{semantic verification} using a \textit{KV-steered Affordance Router} to assess subgoal and action validity without invoking runtime full reasoning.

\item We validate \system{} across simulation and real-world robotic deployments, showing substantial reductions in reasoning latency and redundant computation with only minor impact on task performance.

\end{enumerate}

\section{Background and Related Works}
\label{sec:Background_and_Relworks}

\subsection{On-Device Deployment of ReAct-Based Robotic Policies}
Foundation-model-based ReAct policies can support complex, long-horizon tasks, but their computational demands make on-device deployment challenging. Sequential reasoning and action generation can introduce multi-second per-step latency, which conflicts with the need to continuously process observations arriving at short control intervals~\cite{yin2025deepthinkvla,dai2025actionflow}.
Many existing systems therefore assume workstation-grade compute or remote access~\cite{wen2025diffusionvla,jiang2026vlaperf}. Such offloading can be limited by unstable connectivity and variable communication latency, shifting computation back onto resource-constrained onboard hardware. For example, OpenVLA requires approximately 15GB of VRAM and generates only six complete robot actions per second on an RTX~4090~\cite{kim2024openvla,nvidia4090}, far beyond platforms such as the Jetson Orin NX~\cite{nvidiaorin}. Sustained onboard execution further increases pressure on memory, compute, and energy resources~\cite{mohamed2021energy,lee2025gmt,kang2026foreground}.

\subsection{Efficient LM-Based Robotic Decision Making}

Prior work has reduced the cost of LM-based decision making in several complementary directions. Model-level approaches reduce the computational cost of individual reasoning calls through techniques such as model compression~\cite{yang2025efficientvla}. At the framework level, hierarchical systems separate expensive high-level reasoning from lower-level execution so that a generated plan can guide multiple primitive actions~\cite{song2022llmplanner,lee2026jarvis}. LLM-Planner follows this structure in robot decision making by generating a sequence of high-level subgoals that are subsequently grounded into low-level actions, with replanning triggered when execution fails or progress stalls~\cite{song2022llmplanner}. More recent systems improve other parts of this execution pipeline through multi-rate high- and low-level processing or temporal reuse of previously generated reasoning~\cite{zhang2025hirt,duan2025fastecot}. 

These techniques primarily reduce the cost of generating, executing, or reusing high-level information once available. Our work focuses on a different optimization point within the same hierarchical setting, namely the per-input \textit{reasoning decision} that determines whether the current high-level reasoning remains sufficient or fresh reasoning should be invoked. Improving this decision directly controls how often expensive reasoning enters the execution pipeline while preserving the underlying planner and low-level policy.

\subsection{Navigation and Manipulation Tasks}
Navigation and manipulation capture two complementary forms of robotic interaction with the physical world. Navigation primarily requires scene-level reasoning to select movements, change viewpoints, and respond to evolving surroundings without directly modifying the environment~\cite{shah22lmnav}. Manipulation, in contrast, requires fine-grained reasoning over objects and actions that physically alter the environment~\cite{li2024preciseaffordance}, as exemplified by embodied task benchmarks such as ALFRED~\cite{shridhar2020alfred}. Despite differences, both domains require repeated decisions under changing observations and external environmental effects. Evaluating \system{} across navigation and manipulation therefore allows us to examine whether computationally efficient LM-based decision making generalizes across both high-level spatial reasoning and fine-grained physical interaction.

\section{Motivation}
\label{sec:Preliminary_Study}

To better understand the challenges and motivate our work, we investigate two opportunities for reducing the cost of robotic decision making. First, we examine whether previously generated high-level reasoning can be safely reused through lightweight structural verification. Second, we study whether more ambiguous cases can be resolved using lower-cost semantic verification before invoking full reasoning. Both use ALFRED~\cite{shridhar2020alfred}, a benchmark for long-horizon robotic tasks.

\subsection{Temporal Persistence and Structural Verification}

\begin{figure}[t]
    \centering

    \begin{subfigure}[t]{0.25\linewidth}
        \centering
        \includegraphics[width=\linewidth]{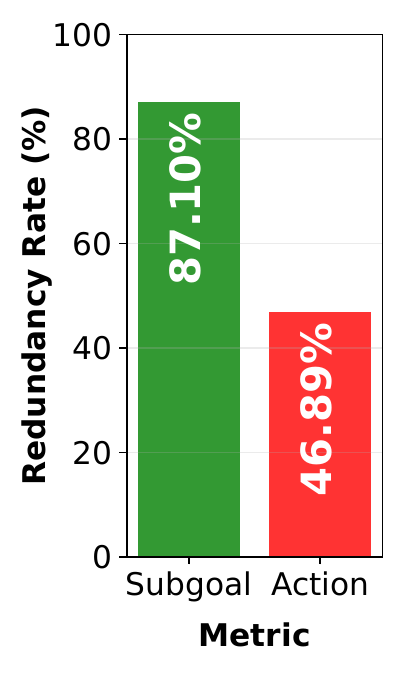}
        \caption{Redundancy rate.}
        \label{fig:redundancy_rate_bar_styled}
    \end{subfigure}
    \hfill
    \begin{subfigure}[t]{0.70\linewidth}
        \centering
        \includegraphics[width=\linewidth]{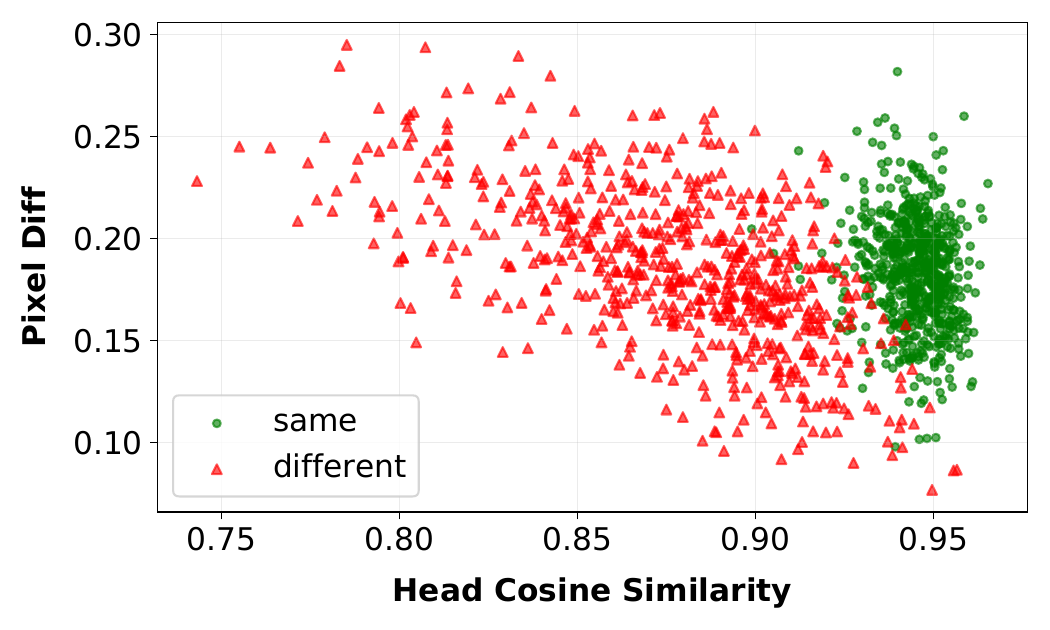}
        \caption{Comparison of pixel-level difference and cosine similarity of the vision encoder head.}
        \label{fig:head_cosine_vs_pixel_diff}
    \end{subfigure}

    \vspace{-1ex}
    \caption{Analysis of redundancy across consecutive inputs.}
    \label{fig:visual_redundancy}
    \vspace{-2ex}
\end{figure}

We first examine whether robotic execution exhibits sufficient temporal redundancy to reuse previous high-level reasoning across consecutive input frames. Figure~\ref{fig:redundancy_rate_bar_styled} quantifies persistence at the subgoal and action levels, defined as the fraction of consecutive steps for which the corresponding output remains unchanged. Both exhibit substantial redundancy, with markedly stronger persistence at the subgoal level. This suggests that high-level decisions evolve more slowly than fine-grained execution, allowing prior reasoning to remain useful across multiple action updates.

We next examine whether such reuse can be determined from consecutive visual observations. We construct controlled ``same'' and ``different'' pairs, where ``same'' preserves the task-relevant scene despite viewpoint changes and ``different'' introduces meaningful scene or object changes. We compare two lightweight signals. Pixel difference captures raw visual change but is sensitive to camera motion and viewpoint shifts. In contrast, \textit{head cosine similarity} compares activations from the same transformer head across two observations, providing a compact measure of scene consistency that is less sensitive to superficial visual changes.

Figure~\ref{fig:head_cosine_vs_pixel_diff} shows that head cosine similarity separates ``same'' and ``different'' observations more clearly than pixel-level difference. The green ``same'' observations concentrate at high similarity on the $x$-axis, while the red ``different'' observations extend toward lower similarity, whereas the two groups overlap along the pixel-difference $y$-axis. This suggests that intermediate visual representations provide a more stable signal of task-relevant scene consistency and motivates their use for the first-stage \textit{structural verification} of \system{}.

\subsection{Low-Cost Semantic Verification}

Structural verification can filter observations that are unlikely to require further reasoning, but visual change alone cannot determine whether the current high-level decision should be revised. Such cases require semantic verification, yet invoking full reasoning for every detected change would diminish the efficiency gained from the first stage.

We therefore examine whether semantic verification can retain much of the benefit of full reasoning at lower computational cost. We consider \textit{KV steering}, which captures internal context associated with full reasoning in a reusable representation prepared offline. Injecting this representation during inference enables reasoning-informed semantic judgments without executing the full multi-step reasoning process. This provides a lightweight path for approximating reasoning-level semantic capability before resorting to full reasoning.

\begin{table}[tb]
\centering
\footnotesize
\setlength{\tabcolsep}{4.5pt}
\renewcommand{\arraystretch}{1.0}
\caption{
Semantic verification on ALFRED action selection.
B and KV denote inference without/with KV steering, while Direct and Reasoning denote direct/full-reasoning inference.
}
\label{tab:kv_priming_main}
\vspace{-1ex}
\begin{tabular}{@{}lcccc@{}}
\toprule
\textbf{Metric} &
\textbf{B-Direct} &
\textbf{B-Reasoning} &
\textbf{KV-Direct} &
\textbf{KV-Reasoning} \\
\midrule
Acc. $\uparrow$ &
21.1\% &
34.5\% &
67.6\% &
\textbf{85.2\%} \\
\bottomrule
\end{tabular}
\vspace{-4ex}
\end{table}

Table~\ref{tab:kv_priming_main} evaluates whether KV steering can provide the intermediate semantic capability required by our cascade. The four settings vary the use of KV steering and full reasoning. Accuracy measures action-selection correctness. The improvement from B-Direct to KV-Direct shows that KV steering can provide substantial semantic information through a lighter inference path without full reasoning. Meanwhile, the additional improvement from KV-Direct to KV-Reasoning shows that harder cases still benefit from deeper reasoning. Together, these results support KV steering as an intermediate.

\section{Method}
\label{sec:System}

Building on the motivated two-stage verification, \system{} augments an existing ReAct pipeline with \textit{System One}, which determines whether the current high-level task decision can be preserved before invoking full reasoning in \textit{System Two}.

\begin{figure*}[!t]
\centering
\includegraphics[width=0.90\linewidth]{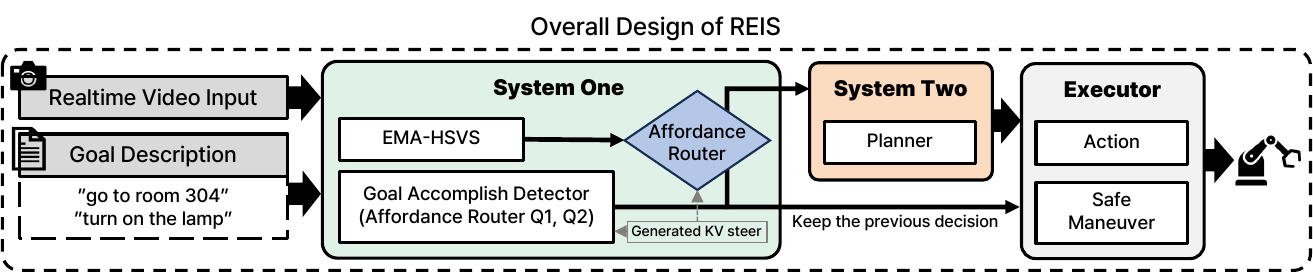}
\caption{Overview of \system{} and the interaction between System One,
System Two, and the Executor.}
\label{fig:system_overview}
\end{figure*}

\subsection{Overall Design and Processing Pipeline}

Figure~\ref{fig:system_overview} shows the complete input-to-action loop of \system{}. Before runtime, \system{} prepares the task-specific components of System One by selecting the transformer heads and similarity threshold for structural verification and constructing the KV steering representations for semantic verification.

At runtime, given a continuous video stream and goal description, \system{} maintains the current high-level task decision as an ongoing subgoal and action. For each new observation, System One first invokes structural verification and follows it with semantic verification only when richer task-level evidence is needed. If the current decision remains valid, full reasoning is skipped and execution continues with the preserved decision. Otherwise, System Two performs full reasoning to update the subgoal and action before execution. The resulting decision is then carried forward as the reference for subsequent observations.

\subsection{System One}

Within this flow, System One resolves the reasoning decision through two stages, structural verification and semantic verification.

\subsubsection{Structural Verification with EMA-HSVS}

The first stage determines whether the task-relevant scene has meaningfully changed and thus the current high-level task decision can be preserved. This verification is repeatedly applied to incoming observations, requiring a lightweight signal that avoids unnecessary triggers from visual variation unrelated to the task.

Inspired by the observation in Sec.~\ref{sec:Preliminary_Study} that selected transformer heads can reliably distinguish task-relevant structural changes from superficial visual variation, \system{} uses these internal representations for structural verification. Before runtime, \system{} uses calibration data and greedy search to select a compact set of informative heads together with the similarity threshold used during execution. At runtime, \system{} computes the cosine similarity between the current and previous head representations to derive the structural similarity signal.

A further challenge is that the robot itself can obscure meaningful changes in the surrounding scene. Since the robot's body or manipulator remains visually similar across observations, its representation can dominate the similarity measurement and make two observations appear more similar even when task-relevant regions have changed. To prevent this effect, \system{} introduces \textit{Self-Mask}, which uses the known camera configuration and robot state to identify image regions occupied by the robot and exclude them from the comparison.

Together, head-selective similarity and Self-Mask form \textit{Ego-Motion Aware Head-Selective Visual Similarity} (EMA-HSVS), the structural verification stage of System One. EMA-HSVS preserves the current high-level task decision when the scene remains structurally consistent and otherwise forwards the observation to semantic verification.

\subsubsection{Semantic Verification with Affordance Router}

Whenever the structural verification stage detects a meaningful scene change, \system{} proceeds to the second stage of System One, semantic verification. However, a structural change in current input does not necessarily mean that the current subgoal or action should be revised. Semantic verification therefore uses richer task-level evidence to determine whether the detected change actually requires updating the current high-level task decision, without immediately invoking full reasoning.

To efficiently make this judgment, \system{} approximates the internal effect of full reasoning without generating the complete reasoning trace. During transformer inference, the KV cache stores key and value representations of previously processed tokens, which provide context for subsequent attention. Therfore, explicit reasoning changes these internal states as the model interprets the task and its progress. KV steering captures this reasoning-induced change and applies it directly to the model's internal state, enabling reasoning-informed semantic judgments without generating reasoning tokens.

To realize this idea, \system{} constructs the steering representations once before runtime. The same affordance inputs are processed through paired executions, one with explicit reasoning before the final decision and the other with direct prediction, and the difference between their KV states captures the internal change associated with reasoning. For each transformer layer $l$ and $X \in \{K,V\}$, we compute the average difference over $N$ pairs as

\vspace{-3ex}
\begin{equation}
S_l^X =
\frac{1}{N}
\sum_{j=1}^{N}
\left(
X_{l,j}^{+} - X_{l,j}^{-}
\right),
\qquad X \in \{K,V\}
\end{equation}

where $(+)$ and $(-)$ denote executions with and without explicit reasoning, respectively. Since both executions address the same affordance judgment and differ primarily in whether reasoning is performed, subtracting their KV states emphasizes internal changes introduced by reasoning while reducing information shared by both execution paths. Averaging this difference across multiple pairs suppresses input-specific variation and yields a more stable reasoning-related steering direction. 

At runtime, the \textit{Affordance Router} processes the current observation, goal description, and task state using the underlying autoregressive model, allowing each judgment to remain conditioned on the current scene and task context rather than relying on a fixed classifier. The prepared steering representations are injected into the corresponding KV states to provide reasoning-informed context during this inference. Instead of generating a full reasoning trace, the router performs compact semantic checks via binary tokens to evaluate high-level task decisions. Specifically, the router determines whether the current subgoal is complete ($Q_1$), whether transition to the next subgoal is valid ($Q_2$), and whether the current action can continue ($Q_3$). If $Q_3$ verifies that the current action can continue, the task decision is safely reused; otherwise, it checks subgoal completion ($Q_1$) and transition readiness ($Q_2$). If none of these lightweight verifications hold, the system escalates to full reasoning. In practice, the first two checks ($Q_1, Q_2$) run continuously or alongside structural checks, while action continuation ($Q_3$) is evaluated when structural verification detects a meaningful scene change.

The \textit{KV-steered Affordance Router} thus provides semantic verification while retaining context-dependent task judgment with minimal decoding. If it verifies that the current high-level task decision can be preserved, execution continues with the existing decision. Otherwise, the observation proceeds to System Two for full reasoning.

\subsection{System Two and Executor}

When System One cannot preserve the current high-level task decision, \system{} returns to the original ReAct path and invokes System Two for full reasoning. Using the current observation, goal description, and execution state, \system{} performs the existing multi-step reasoning process to update the subgoal and action. It then passes either this updated decision or the decision preserved by System One to the Executor for low-level action execution. In this way, \system{} leaves the original reasoning and execution process unchanged while reducing how often full reasoning is invoked.


\section{Evaluation}

\begin{figure*}[!t]
  \centering
  \begin{subfigure}[b]{0.86\textwidth}
    \centering
    \includegraphics[width=\linewidth]{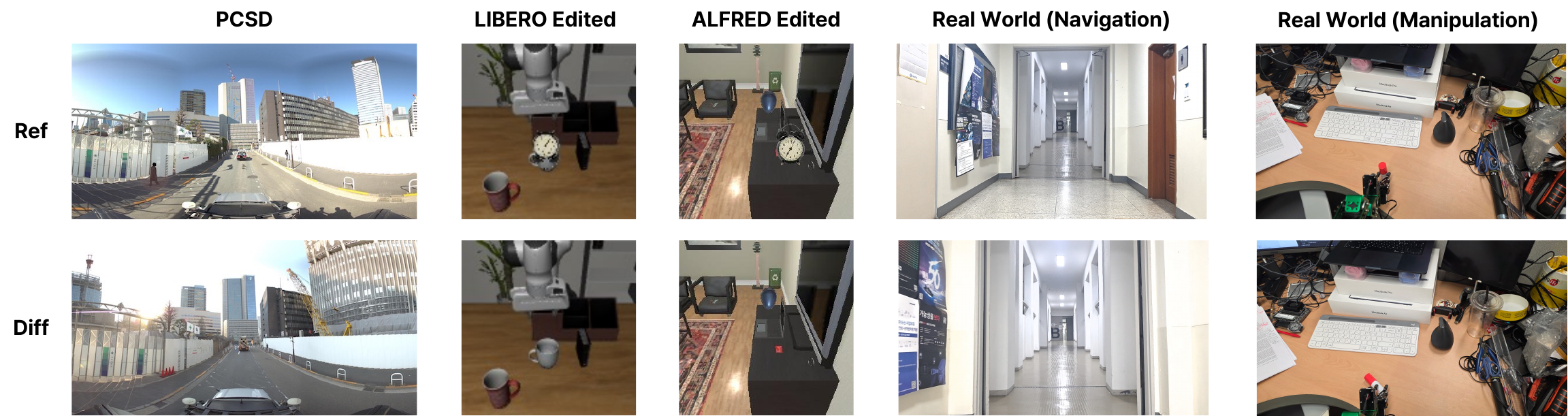}
    \caption{}
    \label{fig:dataset_samples}
  \end{subfigure}
  \hfill
  \begin{subfigure}[b]{0.11\textwidth}
    \centering
    \includegraphics[width=\linewidth]{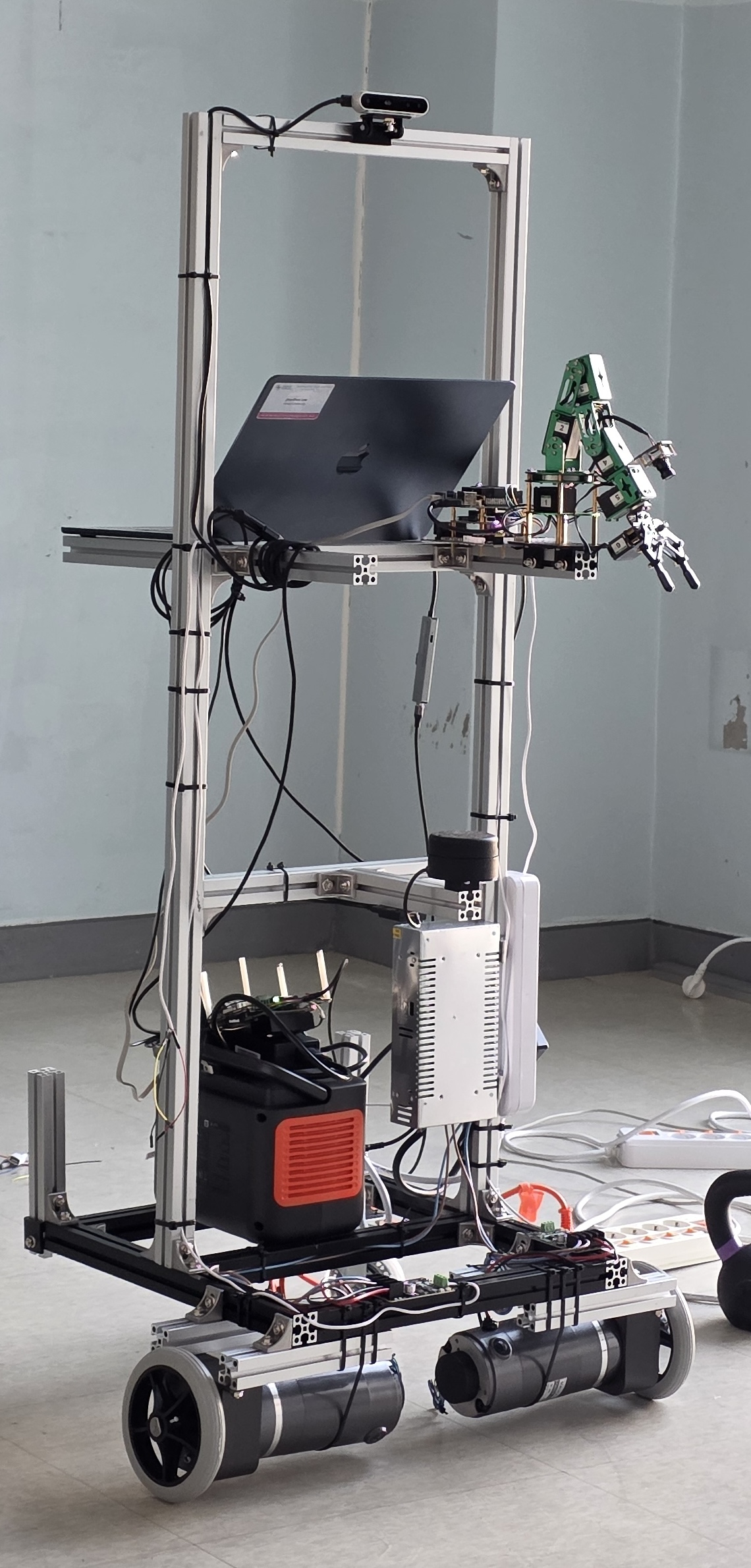}
    \caption{}
    \label{fig:hw_platform}
  \end{subfigure}
  \vspace{-1ex}
  \caption{Experimental setup: (a) Dataset samples used for optimization and evaluation, and (b) Hardware platform.}
  \vspace{-2ex}
  \label{fig:overall_setup}
\end{figure*}

We evaluate \system{} across simulation and real-world robotic tasks to test whether its two-stage reasoning decision reduces full reasoning while preserving high-level task decision quality. We then isolate structural and semantic verification to identify where these gains come from.

\subsection{Evaluation Setup}

Our evaluation covers two complementary embodied domains, namely, navigation with scene-level planning under ego-motion and tabletop manipulation with object-level tracking during local interactions. Figure~\ref{fig:dataset_samples} summarizes the data used across the evaluation, showing reference and changed observations from PSCD~\cite{sakurada2020weakly}, LIBERO~\cite{liu2023libero}, ALFRED~\cite{shridhar2020alfred}, and our RealWorld recordings. These sources provide both scene-level and object-level changes for evaluating whether \system{} responds to task-relevant structural changes rather than superficial visual variations.

All evaluations are conducted on an NVIDIA Jetson Orin NX with 16GB shared memory, where the practical model footprint is limited to approximately 10--12~GB after system overhead. We use compact backbone models, primarily adopting Qwen-3-VL 4B \cite{bai2025qwen3vl} as our representative model since it yielded the most stable results in the primary evaluations. Additionally, we also evaluated Qwen-2.5-VL 3B~\cite{qwen2.5-VL}, Llama-3.2 3B~\cite{meta2024instruct}, and Gemma-3 4B~\cite{gemma_2025} under identical experimental conditions. Every model is served via Hugging Face Transformers engine \cite{wolf2020transformers}.

End-to-end performance is evaluated on ALFRED and through physical deployment on a mobile robot and a single-arm manipulator as Figure~\ref{fig:hw_platform} shows. Navigation and manipulation are evaluated separately so that scene-level transitions and local object interactions are not mixed into the same performance measure. The system is not given an explicit task-type prompt (navigation or manipulation), and mechanical failures such as friction or gripper slippage are excluded from algorithmic decision accuracy. Both ALFRED and the RealWorld deployments are evaluated in a fully closed-loop, on-policy manner under the same definition of these terms.

\subsection{Overall Decision-Making Performance}

\subsubsection{Overall Performance Comparison}

We first test the central system-level hypothesis that \system{} can suppress redundant full reasoning while preserving the high-level task decisions of the underlying planner. All methods receive the same observation stream and task instructions and use the same high-level task decision schema and VLM backbone. We adopt a common hierarchical planner-executor formulation following LLM-Planner, where high-level LM reasoning produces a sequence of subgoals that guides lower-level execution~\cite{song2022llmplanner}. Across all configurations, the high-level planner and low-level execution pipeline remain unchanged, while we vary only how the system determines whether high-level reasoning should be invoked for the current observation. For ALFRED, the backbone is fine-tuned with SFT only to establish a functional planner. Since \system{} leaves the underlying planner unchanged and only determines when full reasoning should be invoked during execution, our evaluation focuses on the resulting efficiency-accuracy trade-off. Specifically, we examine whether \system{} can substantially improve task execution speed while maintaining accuracy close to the underlying hierarchical planner baseline on the seen split. On the unseen split, even this baseline achieves an episode success rate below 10\%, indicating that performance is primarily limited by the underlying planner rather than the reasoning-invocation strategy. Thus, we use the seen split for the end-to-end comparison and separately evaluate the generalization of individual \system{} components.

This controlled comparison isolates the optimization target of \system{}. Rather than modifying how high-level plans are generated or how subgoals are translated into low-level actions, \system{} focuses on the per-input reasoning decision that determines whether the existing high-level reasoning remains sufficient or should be refreshed. As a result, the baseline exploits the same hierarchical planner and executor without the proposed reasoning-decision modules. The Only EMA-HSVS and Only Affordance Router variants further isolate the contribution of the two System One stages.

\begin{table}[t]
\centering
\caption{Overall accuracy and speed performance under a shared hierarchical LM planner-executor. All configurations use the same underlying planner and execution pipeline and differ only in their reasoning-decision mechanisms.}
\label{tab:overall_performance_comparison}
\resizebox{\columnwidth}{!}{%
\begin{tabular}{llcc}
\toprule
\textbf{Dataset} & \textbf{Method} & \textbf{Ep. Acc} & \textbf{Speed} \\
\midrule
\multirow{4}{*}{Real World (Nav.)}
& Hierarchical LM Planner & 0.680 & 0.28 cm/s \\
& Only EMA-HSVS & 0.678 & 4.00 cm/s \\
& Only Affordance Router & 0.643 & 7.14 cm/s \\
& \textbf{\system{} (Qwen-3 VL 4B)} & \textbf{0.641} & \textbf{7.66 cm/s} \\
\midrule
\multirow{2}{*}{Real World (Manip.)}
& Hierarchical LM Planner & 0.760 & 0.71 cm/s \\
& \textbf{\system{} (Qwen-3 VL 4B)} & \textbf{0.729} & \textbf{1.67 cm/s} \\
\midrule
\multirow{4}{*}{ALFRED (Seen Env.)}
& Hierarchical LM Planner & 0.554 & 0.41 f/s \\
& Only EMA-HSVS & 0.532 & 5.61 f/s \\
& Only Affordance Router & 0.551 & 5.14 f/s \\
& \textbf{\system{} (Qwen-3 VL 4B)} & \textbf{0.512} & \textbf{6.36 f/s} \\
\bottomrule
\end{tabular}%
}
\vspace{-3ex}
\end{table}

\begin{figure*}[!t]
\centering
\begin{minipage}[t]{0.61\linewidth}
    \centering
    \includegraphics[width=\linewidth]{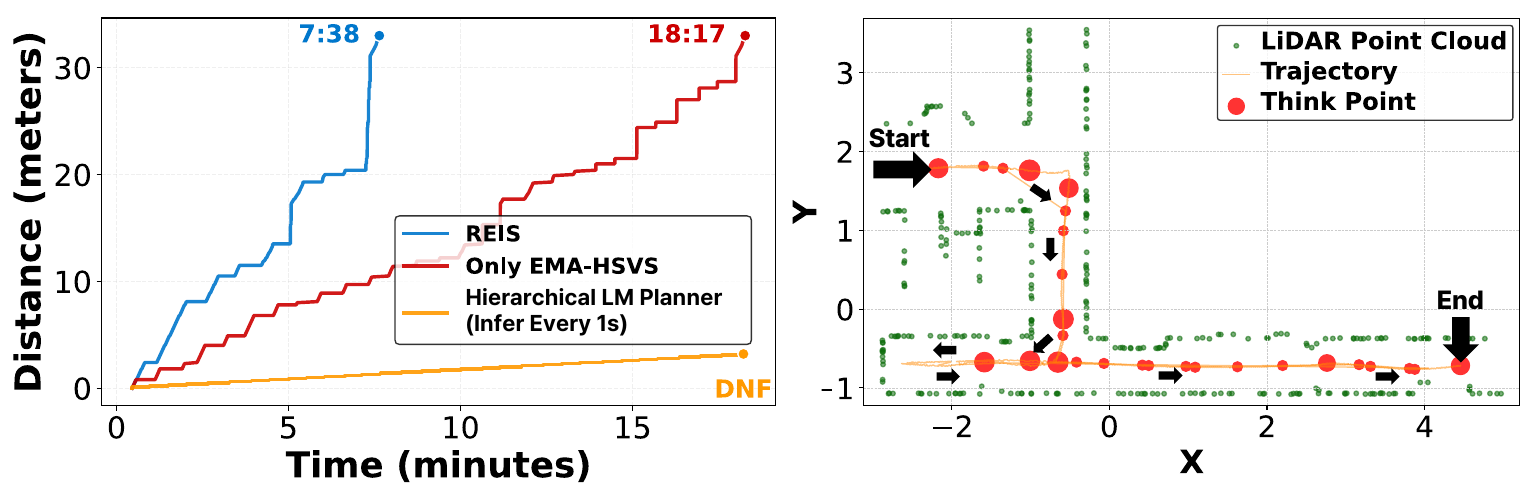}
    \caption{A Sample of Navigation Task (DNF means Did Not Finish)}
    \label{fig:qualitative_analysis_nav}
\end{minipage}%
\hspace{0.02\linewidth}
\begin{minipage}[t]{0.27\linewidth}
    \centering
    \includegraphics[width=\linewidth]{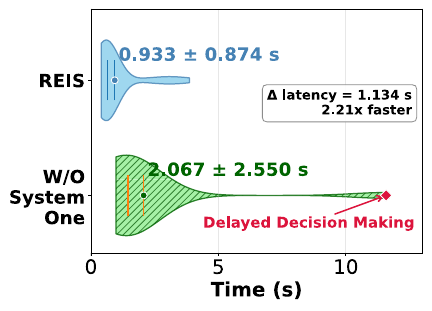}
    \caption{Fail Detect Latency}
    \label{fig:manipulator_failure_dectect_task}
\end{minipage}
\vspace{-2ex}
\end{figure*}

The real-world portion covers navigation across five visually diverse environments and sequential Pick-and-Place manipulation across two tabletop scenes. Table~\ref{tab:overall_performance_comparison} reports episode accuracy together with execution speed. Since \system{} targets reasoning overhead rather than the capability of the underlying planner, execution speed is the primary outcome, while episode accuracy verifies that the speed gain does not come from substantially degrading high-level task decisions. Unless otherwise noted, the primary evaluations use \textbf{Qwen 3-VL 4B} as the backbone model. We additionally evaluate Qwen-2.5 VL 3B, Llama-3.2 3B, and Gemma-3 4B to verify that the behavior of \system{} is not specific to a single backbone.

Across all tasks, \system{} substantially increases execution speed while keeping episode accuracy close to the hierarchical planner baseline. The full system achieves the highest execution speed in every reported setting, showing that the two verification stages provide complementary reductions in reasoning overhead. The gain is most noticeable in the navigation task, where structural verification can bypass repeated reasoning over long stretches of visually and temporally redundant motion. The smaller gain in manipulation is consistent with its use of semantic verification without scene-level structural verification. The ablations further show that EMA-HSVS noticeably removes structural redundancy, while the KV-steered Affordance Router accelerates cases that still require task-level verification. These results isolate the benefit of optimizing when high-level reasoning is invoked while keeping the planner and low-level execution mechanism fixed.

\subsubsection{Navigation Qualitative Analysis}

The aggregate speedup does not alone show where reasoning is avoided in execution, so we inspect a representative real-world navigation trajectory on the Jetson Orin NX. Figure~\ref{fig:qualitative_analysis_nav} complements Table~\ref{tab:overall_performance_comparison} by presenting how progress accumulates over time and where deeper reasoning is invoked along the route. The trajectory contains long intervals in which existing high-level reasoning remains sufficient, while additional verification and full reasoning are concentrated around task-relevant environmental transitions.

The qualitative pattern explains the large navigation gain in Table~\ref{tab:overall_performance_comparison}. \system{} reaches the destination while the hierarchical planner baseline accumulates substantially more reasoning latency, showing that repeated high-level reasoning can directly limit continuous execution on the edge platform. The EMA-HSVS-only variant remains feasible but slower than the complete cascade. This shows that the end-to-end benefit comes from avoiding reasoning at the right observations rather than merely accelerating an individual reasoning call. Furthermore, the left-hand side of Figure~\ref{fig:qualitative_analysis_nav} illustrates that \system{} completes the task in the shortest time due to the reduced number of full reasoning steps. In contrast, the Only EMA-HSVS variant lacks the Affordance Router ($Q_3$), forcing even straightforward decisions to fall back to full reasoning. As a result, we observe that the total execution time increases by $\sim$2.4$\times$ (18m 17s vs. 7m 38s for \system{}).

\subsubsection{Manipulation Failure Detection Analysis}

Manipulation provides a complementary setting in which the scene-level structural stage is disabled since sensitive semantic verification becomes more important for detecting local execution failures.

Figure~\ref{fig:manipulator_failure_dectect_task} compares the failure-detection latency of \system{} with the periodic reference. The important visual pattern is the shift toward shorter response times under \system{}, while the periodic design exhibits a broader delay caused by waiting for a fixed check and then paying full autoregressive reasoning cost. In this experiment, the KV-steered Affordance Router continuously evaluates the most recent incoming frame, dropping intermediate frames to prevent queue accumulation. By performing minimal binary decoding rather than full multi-step generation, it achieves rapid state evaluation. The reported average latency is reduced by 2.21$\times$, showing that the benefit of the reasoning decision also appears as faster responsiveness when a local task state changes.

\subsection{EMA-HSVS for Structural Verification}

Having established the end-to-end benefit, we next isolate the first System One stage. The hypothesis is that structural verification should detect task-relevant scene changes with high sensitivity while avoiding unnecessary (i.e., redundant) semantic verification. EMA-HSVS relies exclusively on structural visual features and does not monitor subgoal completion, leaving such semantic judgments entirely to the downstream router (Affordance Router $Q_1$-$Q_3$). This makes recall important for not missing meaningful changes and precision important for controlling unnecessary downstream checks.

To directly test this property, we construct paired reference and changed observations at both scene and object levels. Figure~\ref{fig:dataset_samples} illustrates this design across PSCD, LIBERO, ALFRED, and RealWorld data. Scene-level pairs capture larger layout changes relevant to navigation, while object-level pairs capture local state and relation changes relevant to manipulation. The calibration set contains 5K pairs for selecting transformer heads, and a separate 1K pairs are used for evaluation. ALFRED calibration instances are drawn from validation-unseen scenes to reduce overfitting to familiar layouts.
We also measure the runtime cost of this first-stage check on the Jetson Orin NX. EMA-HSVS requires an average of 181.42~ms per observation, which should be interpreted relative to the much larger cost of full reasoning that it is designed to avoid.


\begin{table}[t]
\centering
\caption{Comparison of Accuracy, Recall, and Precision at Object and Scene levels (assuming balanced class distribution).}
\label{tab:perf_comparison}
\resizebox{\columnwidth}{!}{%
\begin{tabular}{llccc}
\toprule
\textbf{Level} & \textbf{Method} & \textbf{Acc. (\%)} & \textbf{Recall (\%)} & \textbf{Prec. (\%)} \\
\midrule
\multirow{5}{*}{Object} 
 & Patch Mean Cosine & 60.08 & 80.50 & 57.18 \\
 & CLS Cosine & 78.04 & 69.00 & 84.22 \\
 & CLS L2 & 78.04 & 69.00 & 84.22 \\
 & EMA-HSVS (ours) & 83.83 & 82.67 & 84.63 \\
 & \textbf{EMA-HSVS w/ Self Mask (ours)} & \textbf{98.14} & \textbf{97.17} & \textbf{99.09} \\
\midrule
\multirow{5}{*}{Scene} 
 & Patch Mean Cosine & 71.96 & 89.19 & 66.33 \\
 & CLS Cosine & 79.16 & 88.51 & 74.57 \\
 & CLS L2 & 79.16 & 88.51 & 74.57 \\
 & \textbf{EMA-HSVS (ours)} & \textbf{85.78} & 97.97 & \textbf{78.76} \\
 & \textbf{EMA-HSVS w/ Self Mask (ours)} & 83.29 & \textbf{98.65} & 75.45 \\
\bottomrule
\end{tabular}%
}
\vspace{-5ex}
\end{table}

Table~\ref{tab:perf_comparison} compares generic visual similarity baselines with selected-head EMA-HSVS, with and without Self-Mask, separately for object-level and scene-level changes. Accuracy summarizes overall separation, Recall measures how often task-relevant changes are detected, and Precision measures how often a trigger corresponds to a genuine task-relevant change. The most important columns for the cascade are therefore Recall, which protects against missed changes, and Precision, which controls how often semantic verification is unnecessarily invoked.

The results show that selected transformer heads provide a stronger structural signal than generic patch or CLS representations~\cite{lee2025gmt}. Self-Mask produces a particularly large improvement for object-level changes, where robot-occupied regions can otherwise dominate the comparison, effectively preventing the manipulator arm from being misinterpreted as a meaningful change during local manipulation. At the scene level, Self-Mask slightly increases Recall but reduces Accuracy and Precision relative to EMA-HSVS without masking, revealing a different trade-off rather than a uniform improvement. This occurs because the visible robot body is largely irrelevant to navigation-based layouts, and the masking process inadvertently discards peripheral visual information, subtly degrading the structural judgment. Even with this trade-off, both EMA-HSVS variants maintain very high sensitivity to meaningful changes. These results support the first-stage hypothesis that lightweight structural verification can remove many redundant observations while preserving sensitivity to changes that should reach semantic verification.


\subsection{KV-steered Affordance Router for Semantic Verification}
\label{subsec:affordance_router_evaluation}

Structural verification only establishes that the scene has changed, so we next test whether the second System One stage can determine if that change actually requires revising the current high-level task decision. The hypothesis is that the KV-steered Affordance Router can retain much of the task-level judgment associated with full reasoning while avoiding the complete reasoning trace.

To recap, the router evaluates the current state through three binary affordance questions—checking action continuation ($Q_3$), subgoal completion ($Q_1$), and transition validity ($Q_2$)—to decide whether to preserve the current decision or escalate to System Two for full reasoning.

We compare three inference paths using Qwen3-VL-4B-Instruct. \textbf{Base+Direct} makes the binary decision from the image and prompt without explicit reasoning. \textbf{Base+Reasoning} generates an autoregressive reasoning trace before the decision. \textbf{KV-steered Affordance Router} injects the task-specific steering representation and directly compares the \texttt{yes}/\texttt{no} logits without generating any output tokens. This comparison separates the value of reasoning-informed internal context from the cost of generating the reasoning tokens themselves.

\begin{table}[t]
\centering
\caption{Performance Comparison of System Two Routing}
\label{tab:routing_performance}
\resizebox{\columnwidth}{!}{%
\begin{tabular}{lccccc}
\toprule
\textbf{Method} & \textbf{Acc} & \textbf{Prec} & \textbf{Recall} & \textbf{Avg. Latency} & \textbf{Avg. Tokens} \\
\midrule
Base + Direct & 0.762 & 0.3330 & 0.1390 & 327.56 ms & 0 \\
Base + Reasoning & 0.751 & 0.1000 & 0.2330 & 43.59 s & 290.8 \\
\textbf{Affordance (\system{})} & \textbf{0.913} & \textbf{0.8870} & \textbf{0.8240} & \textbf{406.21 ms} & \textbf{0} \\
\bottomrule
\end{tabular}%
}
\vspace{-3ex}
\end{table}

Table~\ref{tab:routing_performance} reports routing quality in the first four metric columns and inference cost in the final latency and token columns. Accuracy gives the overall routing rate, while Precision, and Recall are more informative about whether the system correctly separates decision preservation from System Two escalation. Therefore, the central comparison is whether the KV-steered router can approach semantic benefits of reasoning while remaining close to the cost of direct prediction.

The KV-steered Affordance Router is the only setting that combines strong routing quality with subsecond edge latency in the reported results. Base+Direct is inexpensive but provides weak separation of the routing states, while Base+Reasoning incurs a much larger latency and suffers from output truncation under the Jetson memory budget, which was directly observed as the cause of lower accuracy by inspecting the raw outputs. The result supports the second-stage hypothesis that reasoning-informed semantic verification can occupy a useful middle point between direct prediction and full reasoning.

The steering representation is prepared only once before runtime, so this cost is separate from the per-observation latency reported in Table~\ref{tab:routing_performance}. The configuration uses $N=10$ contrastive KV pairs, with the preparation time depending strongly on whether generation is performed on the Jetson or an NVIDIA RTX 4080~Ti workstation GPU.

Finally, the low cost of each semantic check makes repeated verification across nearby observations possible. We evaluate this using 150 sequences that transition from a preserved $C_1$ state to System Two escalation, with four subsequent frames sampled at 0.4~s intervals. This experiment tests whether repeated verification across consecutive frames (where consecutive inputs share high similarity) allows the system to cross-check decisions and mitigate potential single-frame errors rather than relying on a single expensive reasoning trace.

Overall, results support the role of the KV-steered Affordance Router as the intermediate semantic verification stage. It preserves context-dependent task judgment at lower cost than full reasoning and leaves unresolved cases to System Two.




\section{Limitations, Future Directions and Conclusion}
\label{sec:Conclusion_n_Discussion}

We presented \system{}, a dual-process framework that minimizes inference latency in reasoning-based robotic policies by partitioning decision-making into an asynchronous EMA-HSVS visual gating mechanism and a fast, KV-steered Affordance Router. 
While our approach substantially reduces computational workload on resource-constrained edge hardware, we see three primary limitations. First, our evaluation heavily leverages benchmarks due to the lack of large-scale real-world data, exploiting physical deployments primarily as feasibility checks. Second, while the VLM backbone was fine-tuned for the ALFRED benchmark, we were unable to apply fine-tuning to the real-world deployments; inherently limiting its performance within existing semantic boundaries. Finally, our current framework is restricted to the high-level function-calling paradigm rather than full end-to-end continuous control~\cite{black2024pi0, chi2023diffusionpolicy}. Future work will extend \system{} to broader embodied architectures, integrating affordance routing into diffusion-based policies to adaptively minimize iterative denoising steps \cite{chi2023diffusionpolicy,wang2024onedp, yu2025d3p} and scaling to VLA pipelines for low-level continuous trajectory generation.


\bibliographystyle{IEEEtran}
\bibliography{references, eis-lab}


\end{document}